# ACUURATE SEGMENTATION OF RETINA NERVE FIBER LAYER IN OCT IMAGES

Mahdi Salarian1, Rashid Ansari1, Justin Wanek2, and Mahnaz Shahidi2
1-Department of Electrical and Computer Engineering, University of Illinois at Chicago, USA
2-Department of Ophthalmology and Visual Sciences, University of Illinois at Chicago, USA

**ABSTRACT**

The quantification of intra-retinal boundaries in the Optical Coherence Tomography (OCT) is a crucial task to study and diagnose neurological and ocular diseases. Since the manual segmentation of layers is usually a time consuming task and relies on the user, an excessive volume of research has been done to do this job automatically and without interference of the user. Although, generally the same procedure is applied to extract all layers, but finding the RNFL is typically more challenging due to the fact that it may vanish in some parts of the eye, especially close to the fovea. To have general software, besides using common methods such as applying the shortest path algorithm on the global gradient of an image, some extra steps have been taken here to narrow the search area for Dijstra's algorithm, especially for the second boundary. The result demonstrates high accuracy in segmenting the RNFL that is really important for the diagnosing Glaucoma.

*Index Terms*— OCT, Fovea, RNFL, ILM, RPE

## 1. INTRODUCTION

The recent progress in imaging technology like the optical coherence tomography (OCT) made it easy for the scientist to use non-invasive methods for the diagnosis and study of diseases in biological tissues. So it is used widely in biomedical applications like Ophthalmology [1, 2]. In fact using high resolution volumetric images from the retina allows researchers to focus on the clinical diagnosis of diseases and effects of each disease on the retinal layers and shape. So each OCT system includes software for segmenting layers manually. This software may be an effective tool to help detect the nerve fiber layers and total retinal thickness, important parameters in the diagnosis procedure. It can also be used for extracting other layers and related features. In spite of their effectiveness, using such software can be a tedious task, in addition to being subjective and influenced by the user. So pseudo or completely automatic algorithms may be more effective. Pseudo automatic algorithms should be handled by experts. Most of recent researches have focused on the automatic segmentation of OCT layers. These methods are based on the digital image processing techniques. However these methods are very useful but they are usually complicated due to existence of illumination variation, speckle noise, blood vessels and so on. Therefore, applying preprocessing steps is necessary. This noise reduction step should be performed in a way to keep the sharpness of the edges. This means that smoothing the images by filters such as low-pass filter could affect the segmentation procedure. Considering this fact, some different approaches have been proposed. Ishikawa [3] used modified mean filters and adaptive thresholding of each A-scan line while [4, 5, 10] applied diffusion filtering to reduce the speckle noise. Although they reached a reasonable performance, these methods are sensitive to intensity inconstancy in each individual layer. This issue reduces the performance when blood vessels are present or images suffer from low contrast. Also, in some images because of special pathologic cases, some layers may vanish which in turn makes it difficult to be distinguished by a simple procedure. This issue usually occurs, for the most part, in the RNFL boundary. A lot of research has been piloted to extract layers but most of them have been limited to huge computational costs. Most of the initial methods were based on intensity variations and edge detection methods such as the Canny-edge detector. [5, 6]. They have faced some problems due to an inconsistency in the intensity. For example, Fabritius proposed a method to detect the RPE and ILM using the intensity variation [1]. Although, these methods can be effective but they have their own weaknesses. So, new researches have been focusing on more sophisticated solutions. Most of these research are based on optimal graph search and graph theories. Segmentation-based methods confirm effective accuracy. For example S. J. Chiu used graph theory and the shortest path algorithm considering ROI for limiting search space for each layer [2]. This study is similar to S. J. Chiu work, while we tried to select ROI in a way that could be applicable on normal cases, as well as, some abnormal ones. We found that with just changing some parts, ILM, RNFL and RPE could be found easily and accurately. In fact, by using the special procedure the RNFL boundary can be detected accurately. This is important when we know finding the RNFL boundary is vital for clinical investigations. This study is based on limiting the search region and the shortest path algorithm. It shows that it is really effective, especially when other algorithms need more steps and show instability in the fovea. Moreover, this method doesn't need volumetric data and can be applied to each B-scan. Having these layers, enables us to measure the nerve fiber layer thickness and average intensity and total retinal thickness along with RPE intensity in lower execution time, which is an important factor in the development of practical applications. Also, this information is necessary for the studying effect of diseases on nerve fiber layer. In the next section, we discuss regular steps in methods based on the shortest path algorithm. In section 3 our proposed algorithm is described and related result is shown in section 4.

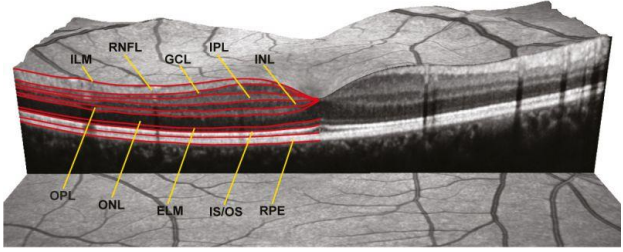

Figure 1    3D image of OCT and its thickness profiles and changes (8)

## 2. RELATED WORK

As mentioned before, our method is based on the shortest path algorithm based on the Adjacency Matrix. Also, we considered ROI for each boundary and used another refining step to address instability issues that might occur in some areas. Here our goal has been to find the RNFL boundary accurately, while detecting other layers needs the same procedure. The shortest path algorithm requires a representation of each image with a graph of nodes and edges. Finding a boundary, in fact, involves going from an initial node to an end node in a way that leads to a less cumulative weight. One of the basic issues in making a graph and its adjacency matrix is the weight assigned to the edges. There are different methods to calculate the edges. For example poor gradient information for each layer considering dark to bright and bright to dark gradients or combining it with the Cany edge detector along with global gradients to compensate each other when one of them suffers from distinctive features. In fact, to find each boundary, one has to run Dijkstra's [9] algorithm considering related ROI. Finding an approximation of the RPE layer is easy, based on the knowledge that this layer usually has the highest contrast and intensity. Based on this approximation, the search space for finding RPE and consequently ILM could be limited. Using Dijkstra's algorithm two layers with the highest contrast, ILM and RPE, can be found easily. The results show that these two boundaries are always present, even in noisy images. Once this boundary has been detected, the next step is image flattering, which transforms images in a way that the shortest path addresses the desired edge. The next step would be to find two complementary gradients of image and calculating two different graph weights regarding dark to light and light to dark edges intensity. Also the weighting scheme such as (1) proposed in [2] is considered:

$$w_{ab} = 2 - (g_a + g_b) \qquad (1)$$

$w_{ab}$ is weight between node $a$ and $b$ and $g_k$ represents gradient in node $k$. Considering different weights also gave us similar results. This means, although, the method for assigning weights is crucial, the instability of some of these schemes may be difficult to solve. For example, Fig. (3) shows the result of the algorithm for two different images. As can be seen, the result for the Fig (3.b) is not accurate. Therefore, ROI is considered as a complimentary step. Instead of working on a comprehensive formula for assigning weights, we have decided to work on another step described in the next sections. The most important challenge has been to find the RNFL boundary. In some images, it is difficult to find the boundary due to the existence of artifacts, especially in the fovea. Consequently, after applying Dijkstra's algorithm on adjacency matrix driven from light to dark gradient image and limiting the space of search below ILM and above RPE, we may come across certain complications. As an example, in figure (3.a ), the RNFL boundary may be thin and very close to ILM or even disappear due to some diseases. Moreover, the next light to dark boundary would be close and so it could be selected. In this study, our goal has been finding the first layer with an acceptable accuracy. It is clear that finding other boundaries would be similar and usually easier than the first one. Here, our proposed method consists of two phases that are described in next section.

## 3. PROPOSED METHOD

### 3.1 Phase 1

This phase is similar to the method used in [2]. After applying a smoothing filter on the whole image and a median filter on each column and considering the threshold, we end up with a binary image. After these steps, a closing operator with the structural element of square with size 2 is applied. This operation is similar to using dilation of an image followed by an erosion to connect some small areas together. In the next step, objects with areas less than 500 pixels are removed. Again, there is another step closing with a structure element of square with size 2 pixel and removing region less than column number (here 1024). Figure (2.a) shows a sample result after this step. The next step is removing the first white region that is supposed to be ILM-RNFL a layer like figure (2.b). The top edge in the resulting image shows the lowest region for searching the RNFL. In fact, ROI for NFL considers some pixels above the first white pixel in this image. The result shown in figure (3) is for a normal case, but sometime there would be some abrupt changes or discontinuities due to vessels or fovea presence in result of the first phase as shown in figure (5). Results show that when there is this effect in result of phase1, the boundary can't be detected properly, especially Assuming that in the presence of fovea, RNFL would be thin and in the worst case, not

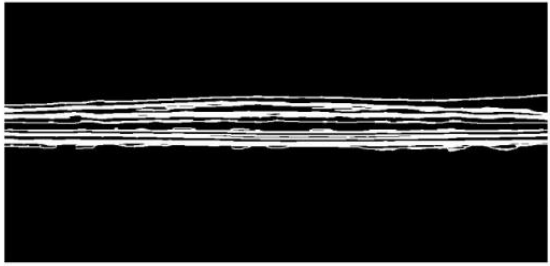 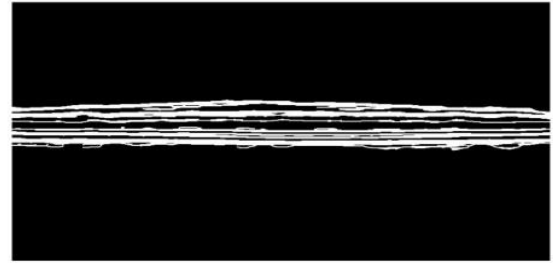

Figure 2    a :applying preprocessing steps,    b: removing first white region detectable even by eyes, we tried to limit the search space by using a copy of the ILM layer, since in this area the RNFL boundary and the ILM could be pretty similar in shape. So in the first step, we found all the discontinuities greater than 3 pixels in resulting images from the past steps. We divided image to 3 parts, left, middle and right. For the middle part, where the fovea is usually present, the outer discontinuity should be selected. For the most cases, there are no discontinuities in the mid-section, while at times one or two discontinuities could be perceived. Nonexistence of discontinuities usually leads to no major issues in the RNFL boundary after applying Dijkstra's algorithm, yet the presence of a discontinuity may cause a copy of the ILM to be used just some pixels below the original ILM as shown in figure (4). The desirable ROI would be between the ILM and this new virtual layer.

The occurrence of discontinuities in the mid-section may owe to existing vessels and could bring about some issues in finding the RNFL layer; this will be discussed further in section 3.2. In fact, half of the average value of the two discontinuities between the ILM and its copy in this area are considered. The following step would be to find the first bright pixel in each column that could be used for the RNFL ROI limitation. We, generally, search for 2 pixels above this edge.

### 3.2 Phase 2

As is mentioned in the prior section, in the low percentage of images discontinuities may occur in the left and right sections and, as a result, it would allow for bigger searching space as shown in the left side of figure (5). As a result, in some cases the shortest path algorithm would go to the next bright to dark edge that is covered by the ROI. Therefore, the first step would be getting rid of this effect. By identifying the positions of the discontinuities, we determined whether each side of the discontinuity was selected correctly. Based on the fact that intensity is lower for the next bright to dark edge, we were able to reach a solution. The first step is having an approximation for intensity value for the first boundary (ILM). After finding the ILM in first phase , the average intensity of columns for 5 pixels below ILM will be calculated. After sorting this vector, values from .7 to .9 of the size of vector are chosen. Higher values are of no interest because at times some types of unusual bright areas occur with noise. For example, with the image of size 1024 100 max values were not considered. Calculating the average of these values would give an approximation of intensity for the first layer intensity that can be used for comparison. Suppose it is called with $I$ the best choice here is finding the approximation using the average of all B scans. But in this study we are trying to develop an algorithm that would be able to be applied on each individual image without having data from other

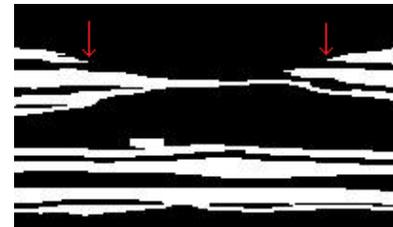

Figure (3): zooming on Fovea and two arrow that show discontinuity in middle section with respect to center of fovea.

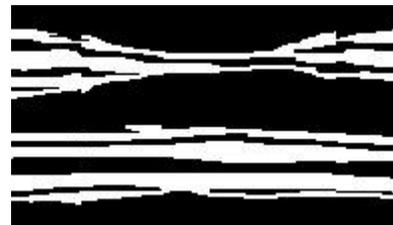

Figure (4): Zoom of figure(7)

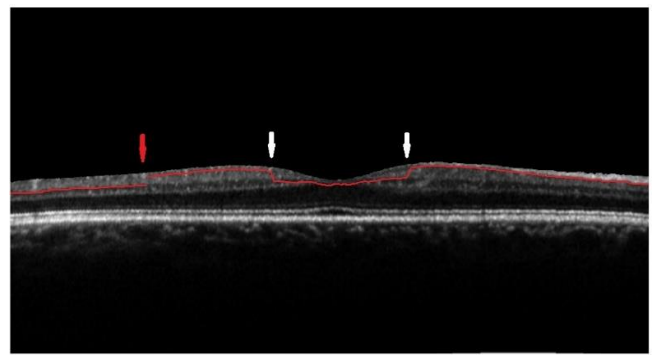

Figure (5): Discontinuity out of middle part (red arrow) and in middle part (white arrow)

B-scans. Now, for each interval containing discontinuity, we attempted to find the number of columns with the depth of 5 pixels and the intensity value of less than $K \times I$. $K$ would be a constant and could be changed but we realized the best value for our work would be 0.9. If the number of columns with a lower intensity than the threshold is less than the given percentage (for example 60 percent), the ROI would not be selected properly and; therefore, should be shifted a few pixels up (3 pixels is enough). Figure (5) shows discontinuities after first step that is just shortest path algorithm. The above procedure could be slightly modified. In some rare situations, especially in pathological cases, images may not have any visible NFL layers in more than half of the images. So, the ROI may be selected wrongly in step 1. By choosing some extra points even without existing discontinuities in the left or right section and applying the procedure described in Phase 2, the ROI would be better selected and have improved results. This procedure was applied on a group of images and the result for one of the worst cases is shown in figure 10.

## 4. CONCLUSION

In short, an automated segmentation algorithm based on the shortest path has been proposed. Knowing this fact, that finding the ILM-NFL boundary is more important for clinical diagnosis and investigation, we focused on finding this layer by finding the proper ROI and correcting this boundary by identifying any instability. We have applied this method to a great number of images including low quality images, as well as, some with pathological eyes; and found it to be accurate. This is a fast method, even when written by MATLAB, and it is supposed to be even faster under C programing. In the future we will be applying other schemes of weighting and use other types of information that can be accessed from the neighbor B-scan to improve our method. This modification can be useful especially for savior pathological cases.

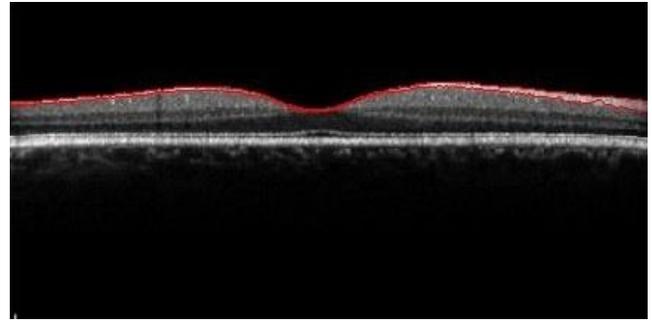

Figure(6):result